\documentclass[conference]{IEEEtran}

\usepackage{times}

\usepackage[numbers]{natbib}
\usepackage{multicol}
\usepackage[bookmarks=true]{hyperref}
\usepackage{graphicx}
\usepackage{psfrag,graphicx,epsfig}
\usepackage{epstopdf}
\usepackage{xspace}
\usepackage{xcolor}
\usepackage{float}
\usepackage{placeins}
\usepackage{multirow}
\usepackage{algorithm}
\usepackage[noend]{algpseudocode}
\algnewcommand{\LineComment}[1]{\State \(\triangleright\) #1}
\usepackage{amsmath}
\usepackage{amssymb}
\newcommand{\timeSet}{\mathcal{T}}

\usepackage{mathtools}
\usepackage{prettyref}
\let\labelindent\relax
\usepackage{enumitem}

\usepackage{pgf,tikz}
\usepackage{nowidow}
\usepackage{lineno}
\usepackage{wrapfig}

\usepackage{siunitx}
\usepackage{color}
\usepackage{flushend}

\usepackage{amsfonts}

\usepackage{amsmath}

\usepackage{amsfonts}
\usepackage{graphicx}
\usepackage{subfig}
\usepackage[bookmarks=true]{hyperref}
\usepackage{duckuments}
\usepackage{booktabs}
\usepackage{float}
\usepackage{amsmath}
\usepackage{multirow}

\begin{document}

\title{Simultaneous Trajectory Optimization and Contact Selection for Multi-Modal Manipulation Planning}

\author{\authorblockN{Mengchao Zhang$^{1}$, Devesh K. Jha$^{2}$, Arvind U. Raghunathan$^{2}$, Kris Hauser$^{1}$}
\authorblockA{$^{1}$ University of Illinois Urbana-Champaign $^{2}$ Mitsubishi Electric Research Laboratories (MERL)}}

\maketitle

\begin{abstract}
Complex dexterous manipulations require switching between prehensile and non-prehensile grasps, and sliding and pivoting the object against the environment. This paper presents a manipulation planner that is able to reason about diverse changes of contacts to discover such plans. It implements  a hybrid approach that performs contact-implicit trajectory optimization for pivoting and sliding manipulation primitives and sampling-based planning to change between manipulation primitives and target object poses. The optimization method, simultaneous trajectory optimization and contact selection (STOCS), introduces an infinite programming framework to dynamically select from contact points and support forces between the object and environment during a manipulation primitive. To sequence manipulation primitives, a sampling-based tree-growing planner uses STOCS to construct a manipulation tree.
We show that by using a powerful trajectory optimizer, the proposed planner can discover multi-modal manipulation trajectories involving grasping, sliding, and pivoting within a few dozen samples. The resulting trajectories are verified to enable a 6 DoF manipulator to manipulate physical objects successfully.
\end{abstract}

\IEEEpeerreviewmaketitle

\section{Introduction}\label{sec:introduction}
\begin{figure}
    \centering
    \includegraphics[width=0.47\textwidth]{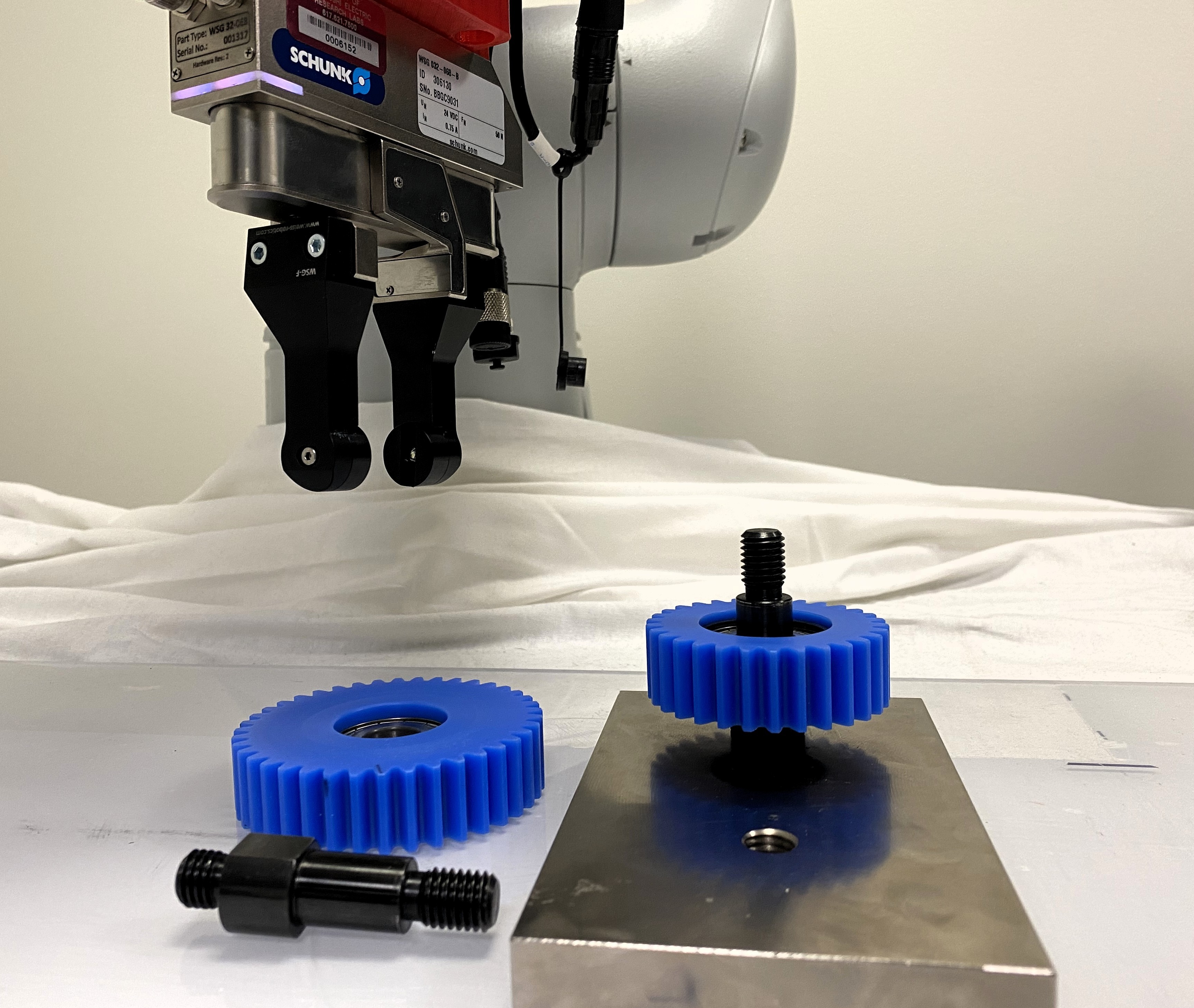} %
    \caption{A robot equipped with a two-fingered gripper may need to re-orient and grasp parts to assemble a gear box. This problem is difficult as the algorithm has to reason about performing a non-prehensile grasp, rotating, and/or sliding before the pick and place action. [Best viewed in color.]}
    \label{fig:exp_setup}
\end{figure}

Humans leverage diverse strategies such as dexterous manipulation, full-body manipulation, and environmental contacts to manipulate a variety of objects, and the robotics field has long attempted to imitate these behaviors~\cite{chavan2020sampling,mason2018toward}. However, it is challenging for optimization-based motion planners to generate these multi-modal behaviors because contacts lead to discontinuous dynamics and the changing numbers of contact points between the robot and the object yield a combinatorial number of possible contact sequences. An example is illustrated in Figure~\ref{fig:exp_setup} where the robot needs to assemble a gear box. Since the blue gear is wider than the gripper's opening, the robot needs to manipulate the gear into an upright pose before it can grasp it. A possible solution is to pivot the gear with a point contact~\cite{9811812} against the external surface before grasping it. However, existing planning and optimization techniques struggle to identify such solutions.

 Contact-implicit trajectory optimization has been explored for its ability to discover complex trajectories for dexterous manipulation \cite{mordatch2012contact} and locomotion problems \cite{mordatch2012discovery}.   
Here, contact is modeled with complementarity constraints between contact forces and relative accelerations, and the optimization is formulated as a mathematical program with complementarity constraints (MPCC)~\cite{raghunathan2022pyrobocop}.
However, it is well known that MPCCs become very challenging to solve as the number of complementarity constraints increase, so past contact-implicit methods were strictly limited to a small handful of potential contact points. This has so far limited their applicability to objects with non-convex and complex shapes. 

This paper introduces the simultaneous trajectory optimization and contact selection (STOCS) algorithm to address the scaling problem in contact-implicit trajectory optimization. It applies an infinite programming (IP) approach to dynamically instantiate possible contact points between the object and environment inside the optimization loop, and hence the resulting MPCCs become far more tractable to solve.  STOCS extends prior work on IP for robot pose optimization~\cite{zhang2021semi} to the trajectory optimization setting. We demonstrate that it can solve for manipulation trajectories involving changes of contact between object and environment.

\begin{figure*}[tbp]
\centering
\setlength\tabcolsep{1pt}
\renewcommand{\arraystretch}{0.0}
\includegraphics[width=0.98\linewidth]{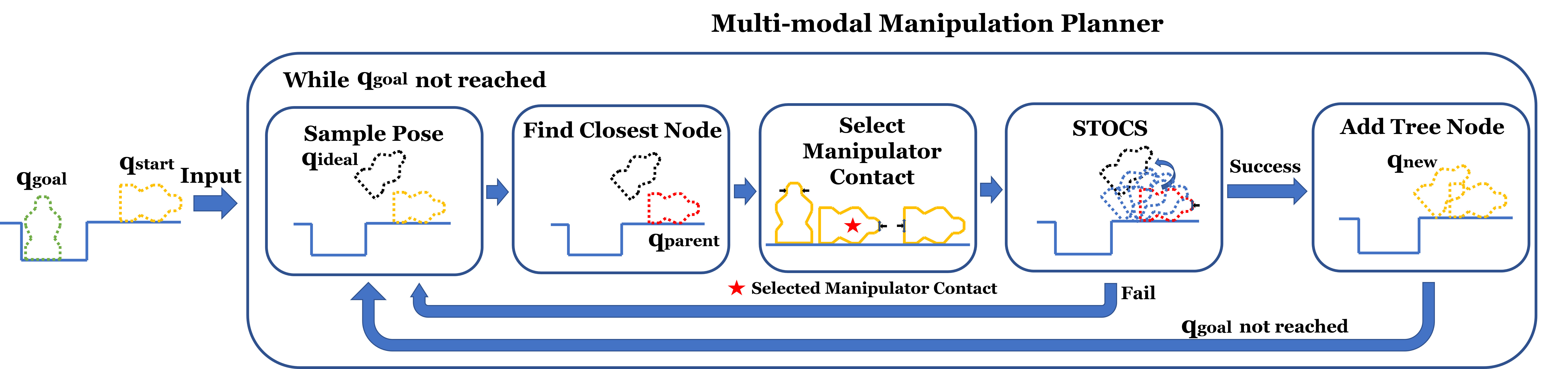}
\caption{Illustrating the workflow of the proposed multi-modal manipulation planner, which could plan for manipulation behaviors consisting a sequence of manipulation modes. Given the start and goal pose of an object as input, the planner  incrementally constructs a manipulation tree through using STOCS to steer the object to reach as far as possible toward randomly-sampled subgoals until the goal pose is reached.\label{fig:multimodal} 
[Best viewed in color.] }


\end{figure*}

Although STOCS is contact-implicit for object-environment contact, it still requires pre-selected robot-object contact state (i.e., a manipulation mode). It is also still a local trajectory optimization method, so its output is sensitive to the specified temporal resolution and initial trajectory.  We address these limitations by introducing a sampling-based planner that uses STOCS to incrementally construct a manipulation tree, which helps it solve for longer-horizon trajectories that change manipulation mode between pushing, pivoting, and grasping.  Each extension of the tree uses STOCS as a local planner to reach as far as possible toward randomly-sampled subgoals. The workflow of the proposed multi-modal manipulation planner is illustrated in Fig.~\ref{fig:multimodal}.

Contributions of this paper include:
\begin{enumerate}
    \item The novel contact-implicit trajectory optimizer STOCS optimizes manipulation trajectories for nonconvex objects and improves computational efficiency by dynamically instantiating nonpenetration, force balance, and complementarity constraints at contact points selected within the optimization loop.
    \item The utility of STOCS is demonstrated as a local planner in a sampling-based planner to extend a tree with sampled manipulation modes toward sampled subgoals.
\end{enumerate}
Our experiments show that STOCS is able to make relatively large jumps through the state space, which helps facilitate rapid progress in planning.  The proposed method is verified on several planar manipulation problems involving pivoting and grasping in simulation, as well as using a physical 6 DoF manipulator.

\section{Related Work}\label{sec:related_work}
Several methods have been proposed to generate manipulation behaviors with changing contacts, and the techniques most related to this paper include contact implicit trajectory optimization, sampling-based motion planning, and task and motion planning. The key challenge is the hybrid nature of contact that splits a solution trajectory into discrete segments, in which contacts are added or removed between segments, but continuous motion within a segment must satisfy multiple continuous constraints, such as force and torque balance, nonpenetration, and complementarity.   


\subsection{Contact Implicit Trajectory Optimization}
Contact-implicit trajectory optimization (CITO) has been studied extensively in dexterous manipulation and legged locomotion literature~\cite{mordatch2012contact, mordatch2012discovery,posa2014direct, manchester2019contact}. It was proposed to overcome limitations of prior trajectory optimization techniques that make use of a pre-defined mode sequence for contact interaction during manipulation/locomotion resulting in a phase-based constrained optimization~\cite{harada2006natural,winkler2018gait}. This allows to cast trajectory optimization as one large non-linear optimization which can be solved to optimize the timings and variables associated with each of the individual modes\cite{chavan2020planar}. However, coming up with such a mode sequence requires contact mode enumeration, which becomes intractable in all but the simplest problems. CITO addresses this problem by modeling all possible points of contact as complementarity constraints, and the trajectory optimization can be cast as a mathematical program with complementarity constraints (MPCC)~\cite{posa2014direct}. CITO can then simultaneously optimize the mode-sequence as well as the contact forces during interaction. 

However, CITO becomes extremely challenging to solve when there are a large number of possible contacts, leading to large numbers of complementarity constraints.  Thus, the main limitation of this approach is that the set of allowable contacts must be predefined and needs to be relatively small. 
In contrast, our approach overcomes this shortcoming by adjusting active contact set simultaneously during trajectory optimization.  

\subsection{Sampling-Based Motion Planning}
Sampling-based motion planning methods such as rapidly exploring random tree (RRT) \cite{lavalle1998rapidly} have proven to be effective for motion planning, and have been used to generate dexterous and nonprehensile manipulation trajectories~\cite{chavan2020sampling,cheng2021contact}. In~\cite{chavan2020sampling}, the T-RRT variant \cite{jaillet2010sampling} is used to plan in-hand manipulation and shows the ability to plan for trajectories which require discrete switch-overs of manipulation contact mode (push location). However, only push motion and only geometry primitives are used in the planning. In \cite{cheng2021contact}, RRT is combined with contact mode enumeration \cite{huang2021efficient} to plan for long-horizon contact-rich manipulation trajectories. Different motion primitives such as pushing, pivoting and grasping could be discovered by this approach, and combination of them can be used to fulfill long-horizon manipulation. 
However, it is difficult to select fruitful contact modes and velocities in random fashion, which leads to very large trees, slow convergence rates, and jerky paths. In contrast, our method achieves generating more efficient trajectories with few nodes in the explored tree by embedding a powerful trajectory optimization as a local planner inside the sampling-based planner.

\subsection{Task and Motion Planning}
Task and Motion Planning (TAMP) solves long-horizon planning problems \cite{garrett2021integrated} by integrating a search over plan skeletons and the satisfaction of constraints over hybrid action parameters. The complexity of planning through contacts is reduced by introducing predefined motion primitives. In \cite{toussaint2018differentiable}, TAMP shows the potential to solve sequential manipulation and tool-use planning problems. However, all objects are assumed to have a sphere-swept convex geometry to make trajectory optimization fast and differentiable, but this assumption is a severe limitation on applicable objects. Also, most of the TAMP methods require expert knowledge and engineering efforts to predefine states and manipulation primitives. On the contrary, our proposed method applies to more general object and environment geometries, and can also discover different motion primitives involving pivoting and sliding within the STOCS optimizer.

\section{Approach}\label{sec:approach}

\begin{figure}
    \centering
    \includegraphics[width=0.47\textwidth]{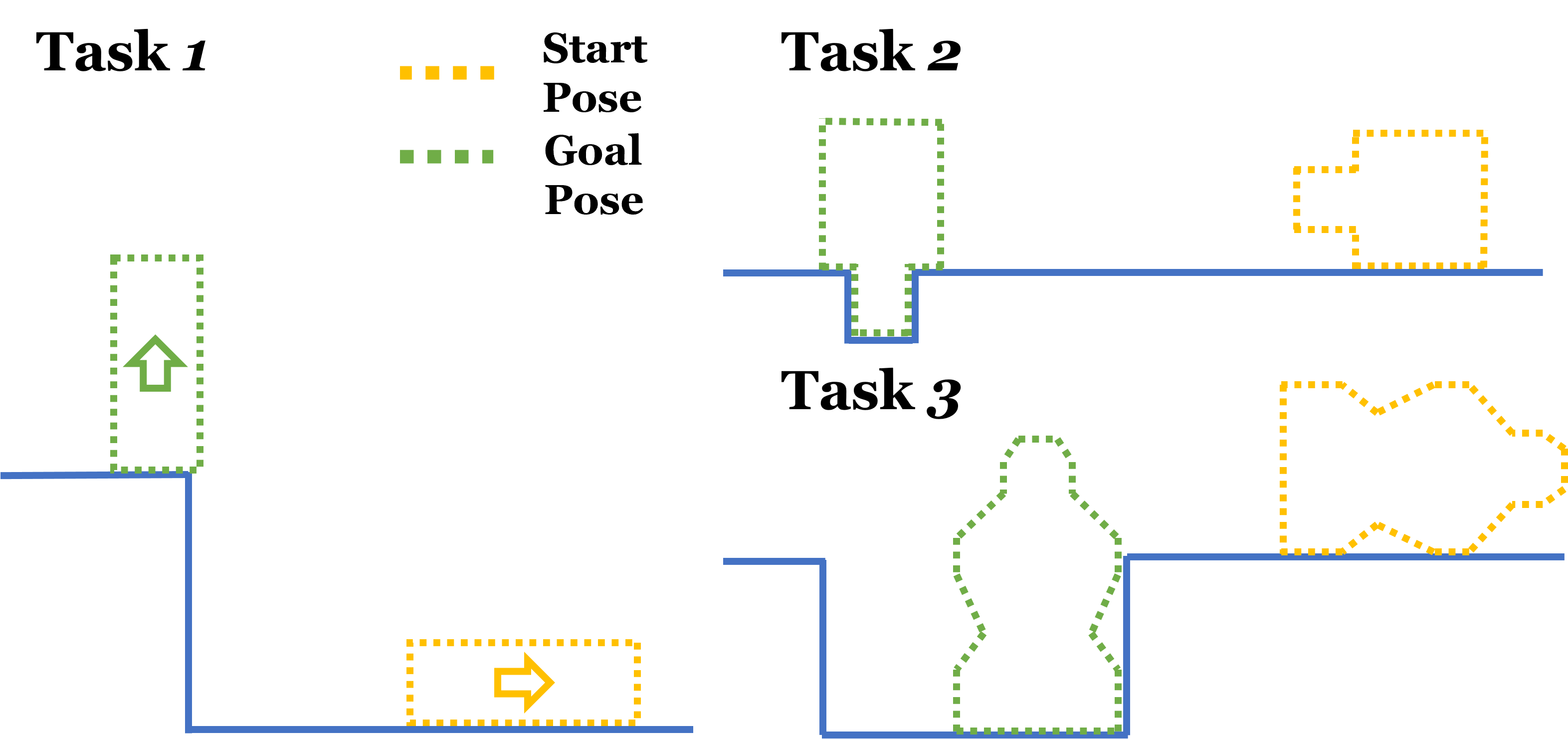} %
    \caption{The start and goal pose of the multi-modal manipulation tasks. In Task 1, the arrow illustrates the orientation of the object. All the objects are not graspable at the start pose. [Best viewed in color.] }
    \label{fig:tasks}
\end{figure}

We wish to plan contact-rich manipulations that may include change of manipulator contact state for arbitrary-shaped object and environment geometries, e.g., reorientation and translation of large parts for assembly, peg in hole, and object packing.  Three example tasks with corresponding initial and goal poses are shown in Fig.~\ref{fig:tasks}.

Our approach consists of two components.  The first is the STOCS algorithm, which solves for object trajectories for a given manipulation mode. The second is a multi-modal sampling-based planner that constructs a manipulation tree to guide the planning towards the goal using STOCS to perform each extension of the tree.  The workflow of the multi-modal planner is illustrated in \prettyref{fig:multimodal}. 

\subsection{Problem Description}\label{subsec:problem_description}

Our method requires the following information as inputs:
\begin{enumerate}
    \item Object initial pose: $q_{init} \in SE(2)$.
    \item Object goal pose: $q_{goal} \in SE(2)$.
    \item Object properties: a rigid body $\mathcal{O}$ whose geometry, mass distribution, and friction coefficients with both the environment $\mu_{env}$ and the manipulator $\mu_{mnp}$ are known.
    \item Environment properties: rigid environment $\mathcal{E}$ whose geometry is known.
    \item Manipulation primitives: all the allowable contact states between the manipulator and the object. 
\end{enumerate}

Our method will output a trajectory $\tau$ which includes the following information at time $t$:
\begin{enumerate}
    \item Object configuration: $q_t$.
    \item Manipulation primitive and manipulator's configuration: $c^{mnp}_t$, and $q^{mnp}_t$.
    \item Manipulation force: $u_t$.
    \item Object-environment contact state $y_t$.
    \item Object-environment force $z_t$.
\end{enumerate}

 We make the following assumptions:
\begin{enumerate}
    \item All objects and environment are rigid.
    \item A set of manipulation primitives are pre-specified as a set of manipulator contact points/surface, specified in object-relative coordinates.
    \item Change of manipulator contact state (i.e., regrasping) is only allowed when the object can stably be supported by the environment alone. 
    \item Quasi-Static: the motion of the manipulated object is slow enough that the forces acting on it are in equilibrium at each instant of time along the trajectory.
\end{enumerate}

\subsection{STOCS Trajectory Optimizer}

\begin{figure*}[tbp]
\centering
\setlength\tabcolsep{1pt}
\renewcommand{\arraystretch}{0.0}
\includegraphics[width=0.7\linewidth]{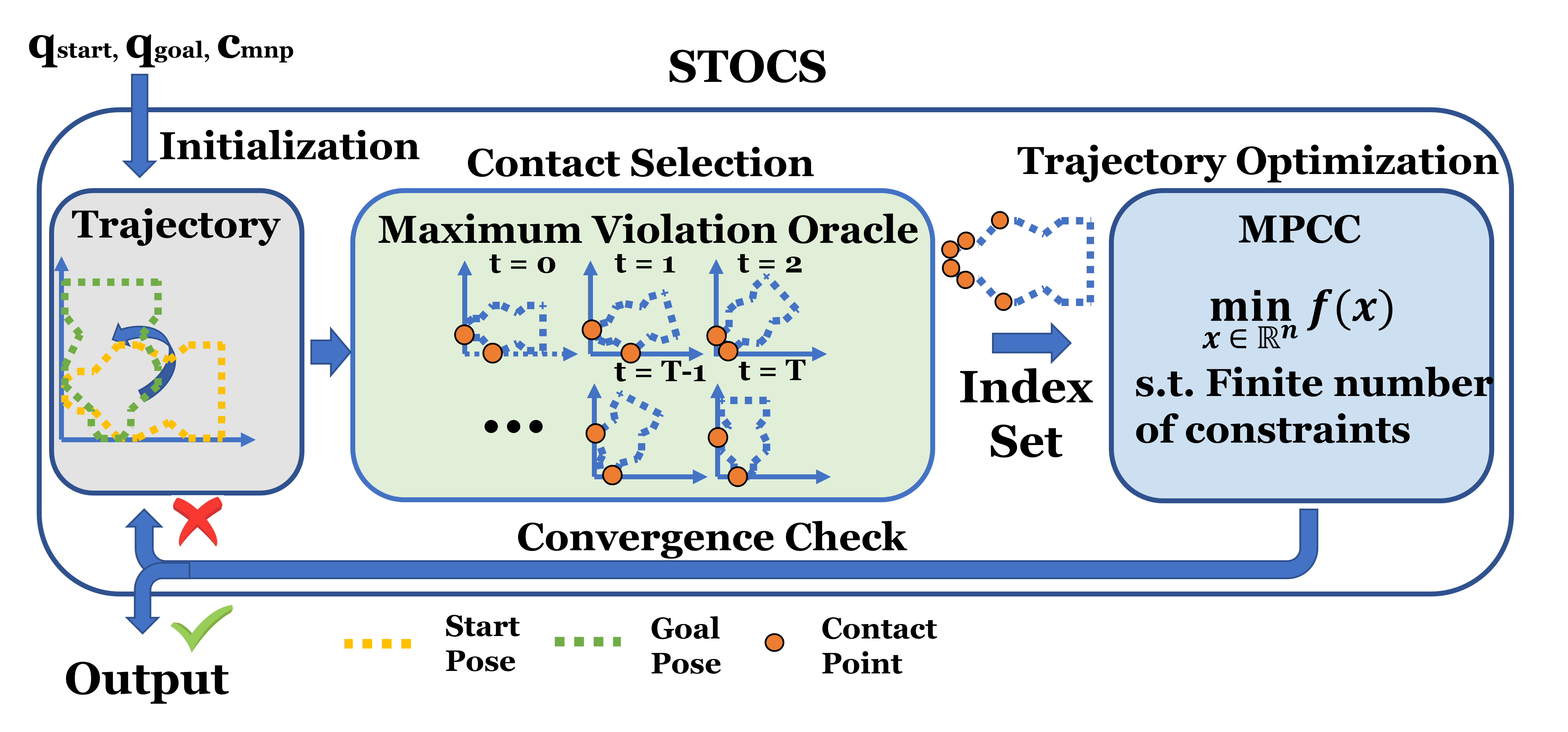}
\caption{\label{fig:flowchart} 
Illustrating the workflow of STOCS. Given the start pose $q_{start}$, goal pose $q_{goal}$, and a manipulator contact state $c_{mnp}$, STOCS iterates between using the Maximum Violation Oracle to instantiate contact points, and solving a finite dimensional MPCC to decide a step direction until the convergence criteria is met. STOCS does not need a predefined contact set and can select the contact points simultaneously while solving trajectory optimization. [Best viewed in color.] }


\end{figure*}

STOCS is a novel CITO algorithm for manipulation that allows multiple changes of contact between the object and environment for a given manipulation mode.  CITO methods formulate contact as complementarity constraints and require solving a MPCC.  STOCS enables the application of MPCC on complex object and environment geometries by embedding the detection of salient contact points and contact times inside the trajectory optimization outer loop. Force variables are introduced for all of these contacts. Each instantiated MPCC iteration has relatively few constraints and is optimized for a handful of inner iterations before new contact points are identified. Force values are maintained from iteration to iteration to warm start the next MPCC. The workflow of STOCS is shown in Fig.~\ref{fig:flowchart}.

\vspace{2mm}
\noindent
\textbf{Semi-infinite programming (SIP), infinite programming (IP) for contact-rich trajectory optimization.}
To illustrate how STOCS works, we start from the SIP/IP problem that STOCS is designed to solve, and explain where the infinitely many optimization variables and constraints come from.
A SIP problem is an optimization problem in finitely many variables $x \in \mathbb{R}^{n}$ on a feasible set described by infinitely many constraints:
\begin{subequations}
\label{1}
\begin{align}
\min_{q\in \mathbb{R}^n} &\; f(q) \\
\text{s.t.} &\; g(q,y) \geq 0 \quad \forall y \in Y 
\end{align}
\end{subequations}
where $g(q,y) \in \mathbb{R}^{m}$ is the constraint function, $y$ denotes the index parameter, and $Y \in \mathbb{R}^{p}$ is its domain. 

In the case of pose optimization with collision constraints, $q$ describes the pose of the object, $y$ is a point on the surface of the object $\mathcal{O}$, where $Y\equiv \partial \mathcal{O}$ denotes that surface of $\mathcal{O}$, and $g(\cdot,\cdot)$ is the distance from a point to the environment. Typically, optimal solutions will be supported by contact at some points, i.e., $g(q^\star,y)=0$ will be met for the optimal solution $q^\star$ at some set of points $y$. 

In the case of trajectory optimization, constraints need to be instantiated in space-time, which means $y = (t,p)$ includes both time $t$ and contact point $p$ on the object's surface. However, to make the resulting optimization problem more stable, in this work, we assume that once a contact point is instantiated, it will be active during the whole trajectory.  

An additional challenge is posed when forces need to be considered as part of the solution to ensure force and moment balance of the object, since force variables need to be introduced to the optimization problem at each point of contact.  We do this by defining $z : Y \rightarrow \mathbb{R}^r$ as an optimization variable. In 2D scenario, for a given contact point $y$, $z=[z^N,z^+,z^-]$ is expressed in a reference frame with $z^N$ normal to the contact surface, and $z^+$, $z^-$ tangent to the contact surface. Given an infinite number of contact points, infinitely many optimization variables will be instantiated, which makes the optimization become an \textbf{IP} problem in which the number of variables and the number of constraints are both possibly infinite \cite{anderson2012infinite}.

In our formulation, each object state along a trajectory must satisfy the following constraints:

\begin{enumerate}[label={\bfseries\arabic*.}, ref=\arabic*]
    \item \textbf{Bound Constraint}: \vspace{-5pt} \begin{equation} q_t\in\mathcal{Q}, \; u_t\in\mathcal{U}, \; z_t(y)\in\mathcal{Z} \; \forall y \in Y \end{equation} \label{2} 
    \vspace{-15pt} 
    \item \textbf{Distance Complementarity Constraint}: \vspace{-5pt} \begin{equation} 0 \leq z_t(y) \perp g(q_t,y) \geq 0 \quad \forall y \in Y \end{equation} \label{3}
    \vspace{-15pt} 
    \item \textbf{Force Inequalities}: \vspace{-5pt} \begin{equation} h(q_t,y,z_t(y)) \geq 0 \quad \forall y \in Y \end{equation}  \label{4}
    \vspace{-15pt} 
    \item \textbf{Control Inequalities}: \vspace{-5pt} \begin{equation} c(q_t,u_t) \geq 0 \end{equation} \label{5}
    \vspace{-15pt} 
    \item \textbf{Integral Constraint}: \vspace{-5pt}\begin{equation} \underbrace{ s_{q,u}(q_t,u_t) + \int_{y \in Y} s_z(q_t,y,z_t(y)) dy}_{\eqqcolon s(q_t,u_t,z_t;Y)} = 0 \end{equation} \label{6}
    \vspace{-1pt} 
\vspace{-10pt}
\end{enumerate}
Eq.~(\ref{3}) ensures that nonzero forces are only exerted at points where objects are in contact, i.e. $z(y) \geq 0$ only if contact is made at $y$, that is $g(q,y) = 0$.  Friction cone constraints are included in the inequalities $h(q,y,z(y)) \geq 0$ in~\eqref{4}.  We constrain the control input in \eqref{5} to make sure the manipulator can only push the object rather than pull the object.  Finally, force and torque balance is expressed in~\eqref{6} as an integral of the force field over the domain $Y$. Here, $s_{q,u}(q,u)$ represents the force and torque applied by gravity and the manipulator, and $s_z(q,y,z(y))$ represents the force and torque applied on an index point $y$ by the contact force $z(y)$. The sum gives the net force and torque experienced by the object, which should be $0$ at quasistatic equilibrium.

We impose these constraints for each state along a trajectory, along with additional constraints that make sure the relative tangential velocity at a contact is zero when the corresponding friction force is inside the friction cone ((\ref{7e}) below). Hence, we formulate the following infinite programming with complementarity constraints trajectory optimization (IPCC-TO) problem denoted as $P(Y)$:
\begin{subequations}
\label{pb:3}
\begin{align}
\min_{q,\dot{q},u,z} &\; 
f(q,\dot{q},u,z)\\
\text{s.t.} & \; q_0=q_{start} \label{7b}\\ 
&\; (2),(3),(5),(6), \dot{q}_t\in\dot{\mathcal{Q}} \;\forall t \in \timeSet \label{7c}\\
&\; q_t - q_{t+1} + \dot{q}_{t}\Delta t = 0 \quad \forall t \in \timeSet \label{7d}\\
&\; 0 \leq v(q_t,\dot{q}_t,y) \perp  h(q_t,y,z_t(y)) \geq 0 \nonumber \\ 
&\; \quad \forall y \in Y, \;\forall t \in \timeSet \label{7e}
\end{align}
\end{subequations}
where $f(q,\dot{q},u,z) \coloneqq \sum_{t\in \timeSet} [f_{q,\dot{q},u}(q_t,\dot{q}_t,u_t) + \int_{y \in Y} f_z(q_t,y,z_t(y)) dy]$,  $\Delta t$ is the time step duration, and $\timeSet = \{0,\ldots,T-1\}$ with $T$ the total number of time steps in the trajectory. For the sake of brevity, we use the notation $q=[q_0, \cdots, q_T]$, $\dot{q}=[\dot{q}_0, \cdots, \dot{q}_{T-1}]$, $u=[u_0, \cdots, u_{T-1}]$, $z=[z_0, \cdots, z_{T-1}]$. With a little abuse of notation, we use $z_t=[z_t(y) \; \forall y \in Y]$ where $z_t(\cdot)$ is the mapping and $z_t$ is a concatenation of all the instantiated variable for all $y\in Y$. 

To solve the IPCC-TO problem, STOCS (Alg.~\ref{alg:1}) includes the following subroutines.

\begin{algorithm}[tbp]

\caption{STOCS}
\label{alg:1}
\begin{small}
\begin{algorithmic}[1]
\Require $q_{start}$, $q_{goal}$, $c_{mnp}$
\State $Y_{0}=[\,]$ \Comment{Initialize empty constraint set}
\State $z_0 \gets \emptyset$ \Comment{Initialize empty force vector}
\State $x_0 \gets $ initialize trajectory($q_{start}$,$q_{goal}$,$c_{mnp}$)
\For{$k=1,\ldots,N^{max}$}

\LineComment{Update constraint set and guessed forces $z_{k}$}
\State Add all points in $Y_{k-1}$ to $Y_{k}$, and initialize their forces in $z_k$ with the corresponding values in $z_{k-1}$ 
\State Call the oracle to add new points to $Y_{k}$, and initialize their corresponding forces in $z_k$ to 0

\LineComment{Solve for step direction}
\State{Set up inner optimization $P_k = P(Y_k)$}
\State Run $S$ steps of an NLP solver on $P_k$, starting from $x_{k-1},z_{k-1}$  
\If{$P_k$ is infeasible}
\Return \textsc{Infeasible}
\Else{}
\State {Set $x^*,z^*$ to its solution, and $\Delta x \gets x^*-x_{k-1}$, $\Delta z \gets z^*-z_{k-1}$}
\State {Do backtracking line search with at most $N_{LS}^{max}$ steps to find optimal step size $\alpha$ such that $\phi(x_{k-1}+\alpha \Delta x,z_{k-1}+\alpha \Delta z;\mu) \leq \phi(x_{k-1},z_{k-1};\mu)$}
\EndIf

\LineComment{Update state and test for convergence}
\State $x_{k} \gets x_{k-1} + \alpha \Delta x$, $z_{k} \gets z_{k-1} + \alpha \Delta z$
\If{Convergence condition is met}
\EndIf
\Return $x_{k}$,$z_{k}$
\EndFor
\Return \textsc{Not\;converged}

\end{algorithmic}
\end{small}
\end{algorithm}

\vspace{2mm}
\noindent
\textbf{Exchange method.}
The IPCC-TO problem $P(Y)$ not only has infinitely many constraints, but also introduces a continuous infinity of variables in $z$. To solve it using numerical methods, we hope that $z$ only is non-zero at a finite number of points. Indeed, if an optimal solution $q^\star$ is supported by a finite subset of index points $\tilde{Y} \in Y$, then it suffices to solve for the values of $z$ at these supporting points, since $z$ should elsewhere be zero. This concept is used in the exchange method to solve SIP problems~\cite{lopez2007semi}, and we extend it to solve IPCC-TO. 

The exchange method progressively instantiates finite index sets $\tilde{Y}$ and their corresponding finite-dimensional MPCCs whose solutions converge toward the true optimum~\cite{reemtsen1998numerical}.  The solving process can be viewed as a bi-level optimization. In the outer loop, index points are selected by an oracle to be added to the index set $\tilde{Y}$, and then in the inner loop, the optimization $P(\tilde{Y})$ is solved. The outer loop will then decide how much should move toward the solution of $P(\tilde{Y})$. Specifically, if we let $(\tilde{x}^*=[q^*,\dot{q}^*,u^*],\tilde{z}^*)$ be the optimal solution to $P(\tilde{Y})$, then as $\tilde{Y}$ grows denser, the iterates of ($\tilde{x}^*$, $\tilde{z}^*$) will eventually approach an optimum of $P(Y)$. 

Given a finite number of instantiated contact points $\tilde{Y} \subset Y$, we can solve a discretized version of the problem which only creates constraints and variables corresponding to $\tilde{Y}$. 
Force variables $z_t$ are instantiated for each index point at each time step, and we replace the distribution $z(y)$ with a set of Dirac impulses: $z_t(y) = \sum_i \delta(y-y_i) z_{t,i}$. Hence, integrals are replaced with sums and we formulate the finite MPCC problem $P(\tilde{Y})$ in the following form:
\begin{subequations}
\begin{align}
\label{pb:4}
\min_{q,\dot{q},u,z} &\; \tilde{f}(q,\dot{q},u,z) \\
\text{s.t.}&\; (2),(3),(5), \dot{q}_t\in {\dot{\mathcal{Q}}} \quad \forall t \in \timeSet\\
&\; (\ref{7b}), (\ref{7d}), (\ref{7e})\\
&\; \tilde{s}(q_t,u_t,z_t;\tilde{Y})=0 \quad \forall t \in \timeSet
\end{align}
\end{subequations}
where $\tilde{f}(q,\dot{q},u,z) \coloneqq \sum_{t=0}^{T-1} [f_{q,\dot{q},u}(q_t,\dot{q}_t,u_t) + \sum_{y \in \tilde{Y}} f_z(q_t,y,z_t(y))] $, and $\tilde{s}(q_t,u_t,z_t;\tilde{Y}) = s_{q,u}(q_t,u_t) + \sum_{y \in \tilde{Y}} s_z(q_t,y,z_t(y)) = 0$. 

\noindent
\textbf{Oracle.}
The key question is how to form the index sets $\tilde{Y}$. A naive approach would sample index points incrementally from $Y$ (e.g., randomly or on a grid), and hopefully, with a sufficiently dense set of points the iterates of solutions will eventually approach an optimum. But this approach is inefficient, as most new index points will not yield active contact forces during the iteration. 

Instead, we use a {\em maximum-violation oracle} that upon each iteration adds the closest / deepest penetrating points between the object and the environment, at each time step along the trajectory. This strategy was used in \cite{hauser2021semi} to avoid collision between robots and environments in trajectory optimization, and our experiments indicate that this approach is successful in trajectory optimization with contact as well. 

\vspace{2mm}
\noindent
\textbf{Merit function for the outer iteration.}
After solving $P(\tilde{Y})$ in an outer iteration, we get a step direction from the current iterate ($\tilde{x}$, $\tilde{z}$) toward ($\tilde{x}^*$, $\tilde{z}^*$). However, due to nonlinearity, the full step may lead to worse constraint violation. To avoid this problem, we perform a line search over the following merit function that balances reducing the objective and reducing the constraint error on the infinite dimensional problem $P(Y)$:

\begin{equation}
\phi(x, z; \mu) = f(x,z) + \mu \|b(x,z)\|_1,
\end{equation}
where $b$ denotes the vector of constraint violations of Problem (8). Also, in SIP for collision geometries, a serious problem is that using existing instantiated index parameters, a step may go too far into areas where the minimum of the inequality $g^*(q)\equiv \min_{y\in Y} g(q,y)$ violates the inequality, and the optimization loses reliability. So we add the max-violation $g^{*-}(x)$ to $b$, in which we denote the negative component of a term as $\cdot^- \equiv \min(\cdot,0)$. 

\vspace{2mm}
\noindent
\textbf{Convergence criteria.}
We denote the index set $\tilde{Y}$ instantiated at the $k^{th}$ outer iteration as $Y_{k}$, the corresponding MPCC as $P_k=P(Y_k)$, and the solved solution as ($x_k$, $z_k$).

The convergence condition is defined as $\alpha \|[\Delta x,\Delta z]\| \leq \epsilon_x\cdot n_{xz}$ \text{and}
$|z_{k}|^T |g(q_{k},Y_{k})| + |v(q_{k},\dot{q}_k,Y_k)|^T |h(q_k,Y_k,z_k)|\leq \epsilon_{gap} \cdot n_{cc}$ \text{and}
$|s(x_{k},z_{k},Y_{k})| \leq \epsilon_s \cdot T$ \text{and}
$\sum_t g_{k}^{-*}(x_{k,t}) < \epsilon_{p}\cdot T$, where $n_{xz}$ is the dimension of the optimization variable and $n_{cc}$ is the number of complementarity constraints, $\epsilon_x$ is the step size tolerance, $\epsilon_{gap}$ is the complementarity gap tolerance, $\epsilon_{s}$ is the balance tolerance, and $\epsilon_{p}$ is the penetration tolerance. With a little abuse of notation, $g(x_{k},Y_{k})$ is the concatenation of the function value of all the points in $Y_k$, and similar for $v(q_{k},\dot{q}_k,Y_k)$ and $h(q_k,Y_k,z_k)$.

\subsection{Multi-Modal Manipulation Planner}

The multi-modal manipulation planner uses sampling to enable robot-object contact state switches, while STOCS is used to optimize changes of contact between the object and environment.  Alg. 2 presents the proposed planner, which combines STOCS with a T-RRT~\cite{jaillet2010sampling} approach to guide tree expansion toward the goal. 

\begin{algorithm}[]

\caption{Multi-modal Manipulation Planner}
\label{alg:2}
\begin{small}
\begin{algorithmic}[1]
\Statex \textbf{Input} $q_{init}$, $q_{goal}$
\Statex \textbf{Output} tree $\mathcal{T}$
\State $\mathcal{T} \gets$ initialize tree($q_{init}$)
\While{$q_{goal} \; \notin \mathcal{T}$}
    \State $q_{ideal} \gets$ sample random configuration($\mathcal{C}$)
    \State $q_{parent} \gets$ find nearest neighbor($\mathcal{T}$, $q_{ideal}$)
    \If{transition test($q_{parent}$, $q_{ideal}$, $q_{goal}$)}
        \State $stable \gets$ stability test($q_{parent}$)
        \If{$stable$}
            \State $c^{mnp} \gets$ sample mnp contact state($q_{parent}$)
        \Else{}
            \State $c^{mnp} \gets$ parent mnp contact state($q_{parent}$)
        \EndIf
        \State $q_{new} \gets$ STOCS($q_{parent}$, $q_{ideal}$, $c_{mnp}$)
        \If{$q_{new} \neq$ null}
            \State add node $q_{new}$ to $\mathcal{T}$
            \State add edge $q_{parent} \rightarrow q_{new}$ to $\mathcal{T}$
        \EndIf
    \EndIf

\EndWhile

\end{algorithmic}
\end{small}
\end{algorithm}

The \textbf{sample random configuration} function has a probability $p_1$ of returning a random sample $q_{ideal}$ from the configuration space $\mathcal{C}$, a probability $p_2$ of returning a random sample whose rotation angle is sampled from the angles of all the possible stable poses of the object on a plane, and a probability $1-p_1-p_2$ of returning the goal configuration $q_{goal}$. Without $p_2$, the probability of a stable pose to be sampled is $0$, and the switch of manipulation contact state will never be triggered. The \textbf{find nearest neighbor} function returns the nearest neighbor of $q_{ideal}$ in the tree $\mathcal{T}$ using the weighted $SE(2)$ metric defined as:

\begin{equation}
    dist(q_1,q_2)=\sqrt{w_1 \cdot (d_x^2+ d_y^2) +w_2 \cdot d_{\theta}^2} 
\end{equation}
where $d_x = q_1^{(x)}-q_2^{(x)}$, $d_y=q_1^{(y)}-q_2^{(y)}$, $d_{\theta} = min(|q_1^{(\theta)}-q_2^{(\theta)}|,2\pi-|q_1^{(\theta)}-q_2^{(\theta)}|)$, and $w_1$ and $w_2$ are the weights that balance the importance between translation and rotation.

The $\textbf{transition test}$ function follows T-RRT~\cite{jaillet2010sampling} to decide whether to accept to propagate the tree from $q_{parent}$ towards the newly sampled configuration $q_{ideal}$ or not. This loosely guides the propagation of the tree toward the goal while still allowing the tree to steer away from the goal with lower probability. We define the cost $C_q$ of a configuration $q$ as $dist(q,q_{goal})$, and the transition test is done by comparing the cost of $q_{parent}$ and $q_{ideal}$.  Define $\Delta C = \frac{C_{q_{ideal}}-C_{q_{parent}}}{dist(q_{parent},q_{ideal})}$ as the normalized change in cost. The new sample will be discarded if $C_{ideal}$ exceeds a maximum bound $C_{max}$, and the new sample will be accepted if $\Delta C \leq 0$. But if $\Delta C > 0$, then $q_{ideal}$ is accepted with probability
\begin{equation}
    p(q_{parent},q_{ideal})=exp\left(\frac{\Delta C_{(q_{parent},q_{ideal})}}{KT}\right),
\end{equation}
where $K$ is a normalization factor defined as the average of $C_{q_{ideal}}$ and $C_{q_{parent}}$, and $T$ is the temperature parameter that is used to control the difficulty level of transition tests. $T$ is adaptively tuned during sampling as in T-RRT. 

The $\textbf{stability test}$ function checks if the object $\mathcal{O}$ is stable at the input configuration in the environment $\mathcal{E}$ without any manipulation force. This is accomplished by solving an IPCC problem. If $q_{parent}$ is stable, then the $\textbf{sample mnp contact state}$ function will randomly select an allowable manipulator contact state $c^{mnp}$ at this configuration.  If $q_{parent}$ is unstable, then the manipulation contact state will be inherited from $q_{parent}$.

\begin{figure}
    \centering
    \includegraphics[width=0.47\textwidth]{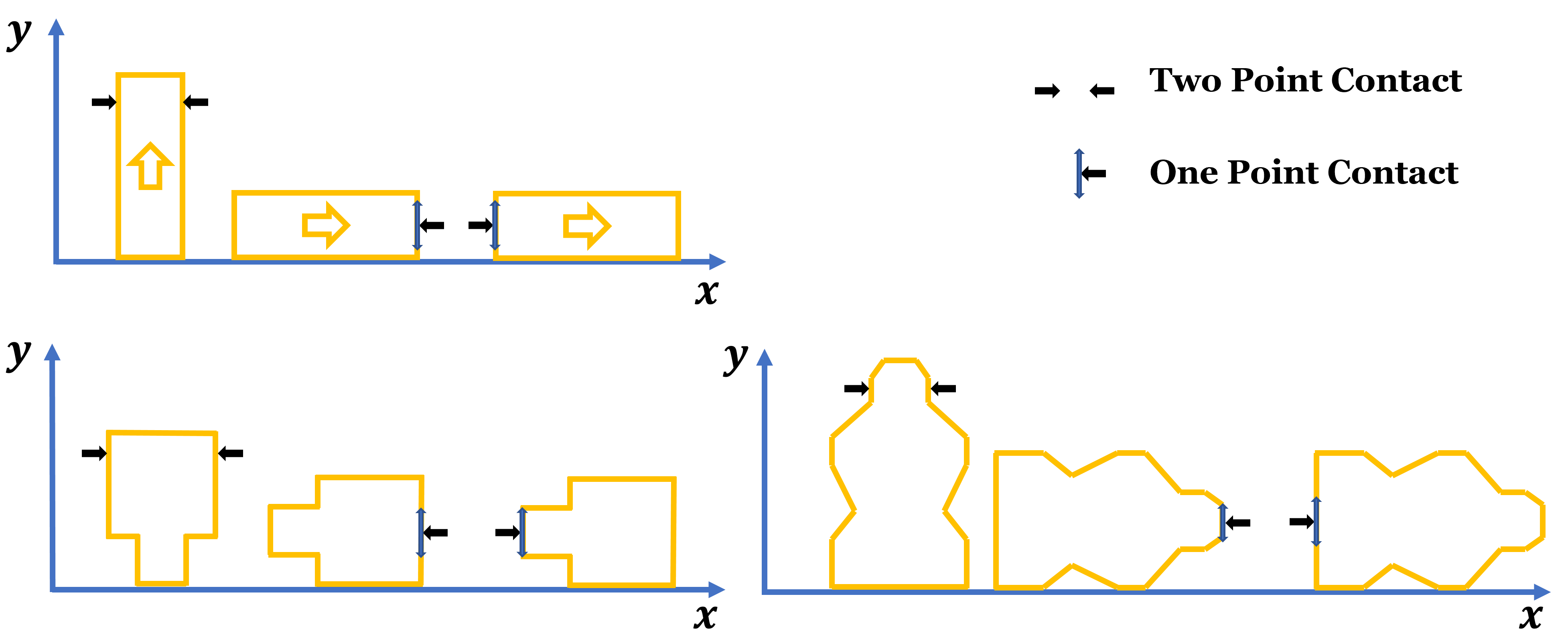} %
    \caption{Allowable manipulator contact states for objects used in the experiments. One-point contact allows the manipulator to slide on a designated surface of the object, and two-point contact is a fixed contact location relative to the object's frame. [Best viewed in color.] } 
    \label{fig:modes}
\end{figure}

The permissible manipulation modes may be defined in a problem-dependent manner to reflect the manipulation primitives available to the robot. The modes used in our experiments are illustrated in Fig.~\ref{fig:modes}. Two-point contact is only possible when the object is at certain upright rotation angles, and one-point contact is permissible at surfaces not in contact with the environment.

For each extension of the tree, STOCS is configured with an objective function $\tilde{f}(q,\dot{q},u,z)= W \sum_{t} dist(q_{t},q_{goal})^2$ to guide the object toward $q_{goal}$ as close as possible. When a stable angle of the object is sampled by \textbf{sample random configuration}, we set $w_1=0$ and $w_2=1$ to disregard the translation components in the objective function, and STOCS will try to steer the object to the target angle. Otherwise, we use $w_1=w_2=1$. $W=5$ is used in the experiments.

\section{Experimental Results}\label{sec:experiment}
We evaluate STOCS and the proposed multi-modal manipulation planner by performing several numerical experiments and some physical experiments. The proposed methods are implemented in Python using the PYROBOCOP framework \cite{raghunathan2022pyrobocop}, which uses the IPOPT solver for optimizations \cite{dikin1967iterative}. All experiments were run on a single core of a 3.6 GHz AMD Ryzen 7 processor with 64 GB RAM. 

\subsection{Experiments on STOCS}


\begin{figure}[tbp]
\centering

\renewcommand{\arraystretch}{0.0}
\includegraphics[width=0.95\linewidth]{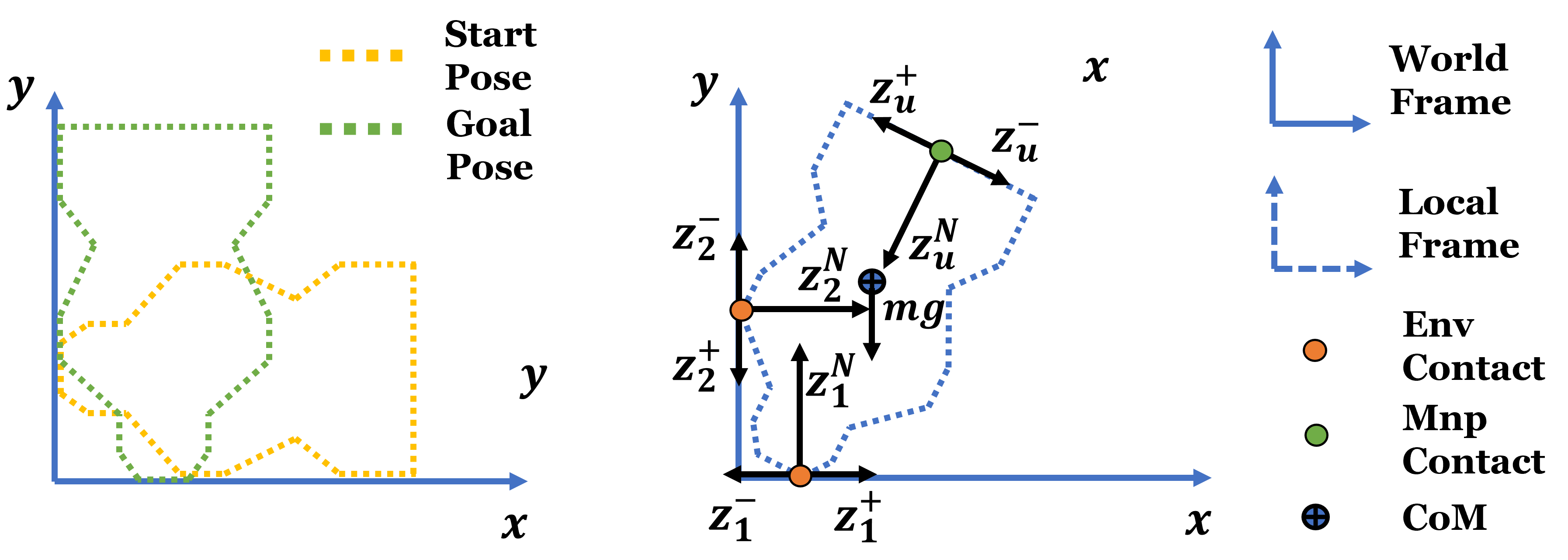} \\
\caption{\label{fig:pivot} Left: start and goal pose of a pivoting task. Right: free-body diagram of the object-robot-environment contact during the pivoting. [Best viewed in color.] }
\vspace{-10pt}
\end{figure}

First, we compare STOCS with vanilla MPCC to evaluate the efficacy of dynamic contact selection.  We finely discretize the object geometries to better illustrate the advantages of our method. Vanilla MPCC involves adding all index points in $Y$ to an MPCC problem without selection, resulting in a larger optimization problem than $P_k$ in STOCS. Both optimization formulations are tested on the pivoting task illustrated in Fig.~\ref{fig:pivot}
. Parameters used in the experiments include:
\begin{itemize}
    \item \textbf{Manipulation mode}: one point of robot-object contact.
    \item \textbf{Physical parameters}: Object mass $m=0.1$ kg, environment friction coefficient $\mu_{env}=0.3$, manipulator friction coefficient $\mu_{mnp}=0.7$.
    \item \textbf{Algorithm parameters}: $N^{max}=100$, $\epsilon_x=\epsilon_{gap}=\epsilon_{s}=\epsilon_{p}=1e^{-4}$, $S=min(30+10 k,200)$, $T=20$ and $\Delta t=0.1$.
\end{itemize}


The results are presented in Table~\ref{tab:1}. We observe that STOCS can be around one to two orders of magnitude faster than MPCC, and can solve problems that MPCC cannot solve due to limitation of computational resources. STOCS selects only a small amount of points from the total number of points in the objects’ representation on average, which greatly decreases the dimension of the instantiated optimization problem and reduces solve time. 

\begin{table}[tbp]
\centering
\renewcommand{\arraystretch}{1.2}
\setlength\tabcolsep{4pt}
\begin{tabular}{@{}lp{1.4cm}llll@{}}
\toprule
\multicolumn{2}{c}{}                                            & \multicolumn{3}{c}{{\bf STOCS}}                                                                                                                                 & {\bf MPCC}     \\ \cline{3-5}
Object  & \# Points &  Time (s) & Outer iters. & Index points & Time (s) \\ \midrule
Box     & 104      &   47.6     & 8                                                           & 2.00                                                                              &    6672.8      \\   
Peg     & 104                          & 223.5    & 13                                                          & 3.23                                                                             &   8573.1       \\ 
Mustard & 247                        & 402.2    & 13                                                          & 4.85                                                                             & OoM      \\ \bottomrule
\end{tabular}
\caption{Numerical optimization results of STOCS and MPCC on the pivoting task. Number of points in the object's representation (\# Point), solve time (Time), outer loop iteration number (Outer iters), and average active index points for each iteration (Index points) are reported in the table. OoM means out of memory. \label{tab:1}}
\end{table}

\begin{figure}[tbp]
    \centering
    \includegraphics[width=0.47\textwidth]{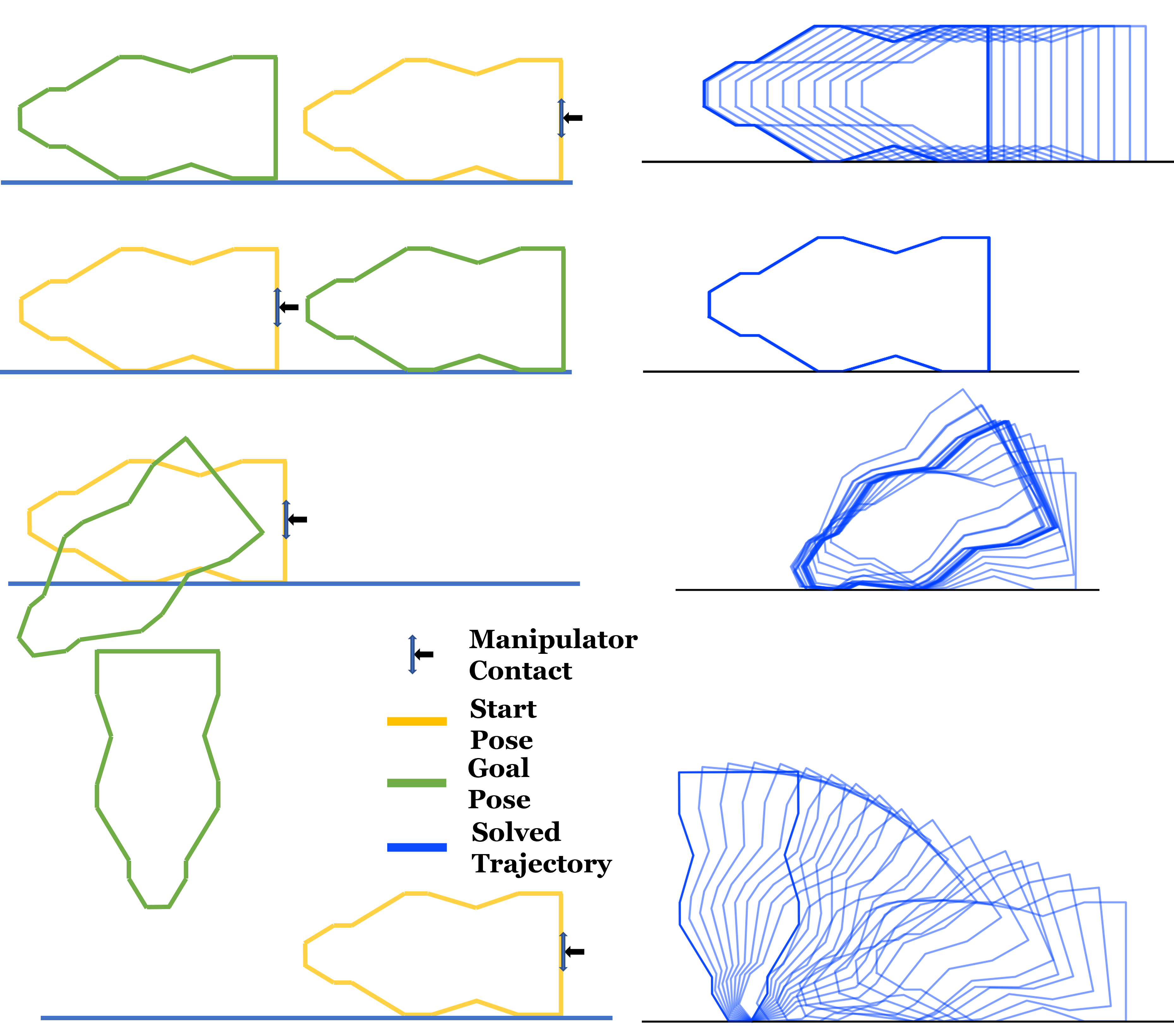}
    \caption{Start pose, goal pose, and selected manipulator contact state are shown on the left side, and the corresponding trajectories solved by STOCS are shown on the right. STOCS steers the object to the goal pose as close as possible while satisfying all the constraints imposed by the selected manipulator contact state and the environment. [Best viewed in color.]} 
    \label{fig:stocs_examples}
\end{figure}

Next, we test STOCS with different manipulator modes and goal poses, including goals that are infeasible or unreachable. As shown in Fig.~\ref{fig:stocs_examples}
, STOCS steers the object to the goal pose as close as possible while satisfying all the constraints imposed by the selected manipulator contact state and the environment.

\subsection{Experiments on Multi-modal Manipulation Planner}


Next we test the proposed multi-modal manipulation planner on tasks requiring one or more changes of manipulation mode. Parameters used in the experiments include:
\begin{itemize}
    \item \textbf{Physical parameters}: Object mass $m=0.1$ kg, environment friction coefficient $\mu_{env}=1.0$, manipulator friction coefficient $\mu_{mnp}=1.0$. 
    \item \textbf{STOCS parameters}: $N^{max}=10$, $\epsilon_x=\epsilon_{gap}=\epsilon_{s}=\epsilon_{p}=1e^{-4}$, $S=min(30+10*k,200)$, $T=5$ and $\Delta t=0.1$.
    \item \textbf{Multi-modal planner parameters}: $C_{max}=2$. Runs are terminated after a maximum of 500 extensions.
\end{itemize}
All experiments in this section are evaluated under $10$ different random seeds.

\begin{figure}
    \centering
    \includegraphics[width=0.47\textwidth]{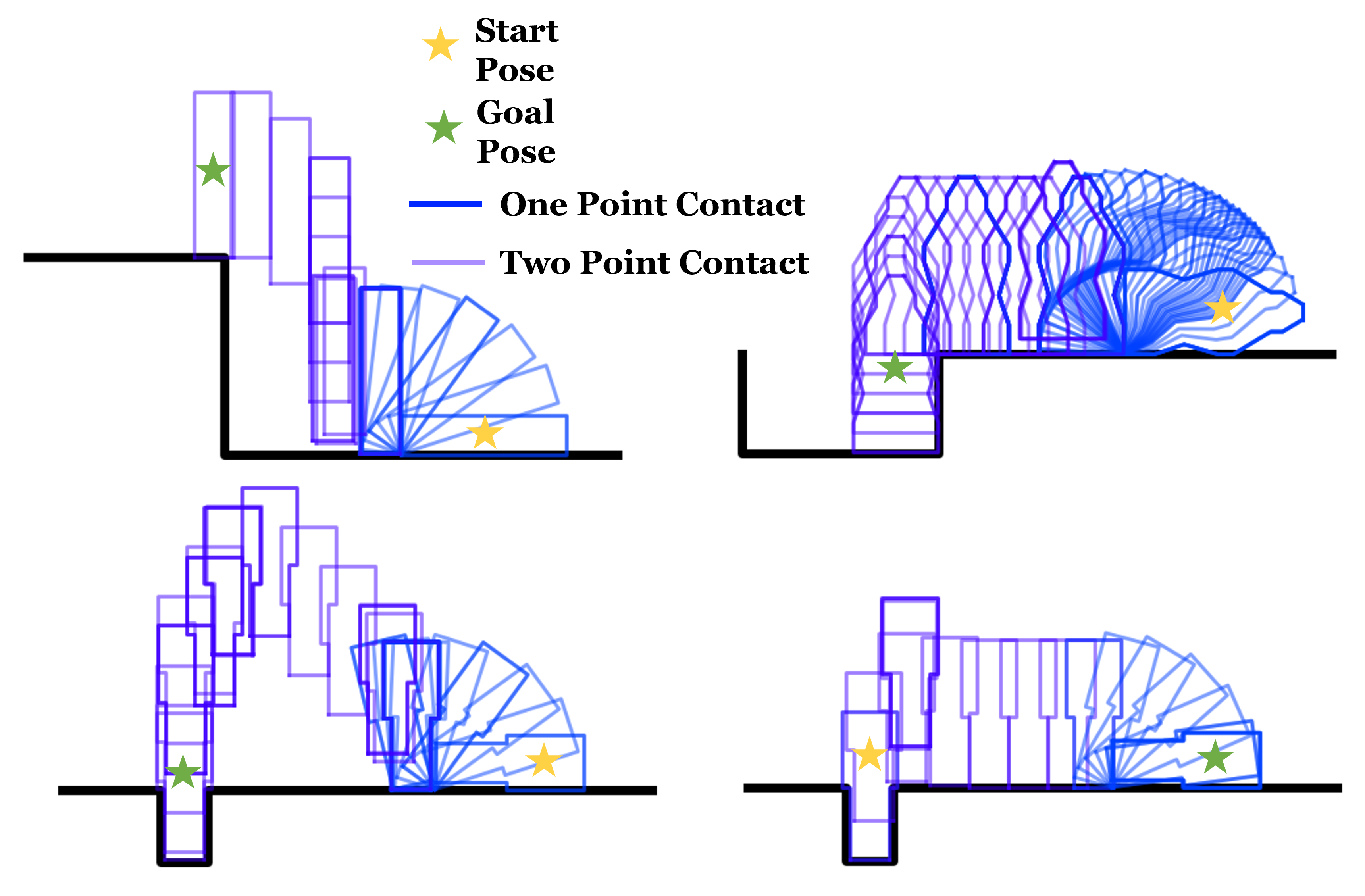}
    \makebox[1.0\linewidth]{\footnotesize (a) Trajectories}
    \includegraphics[width=0.47\textwidth]{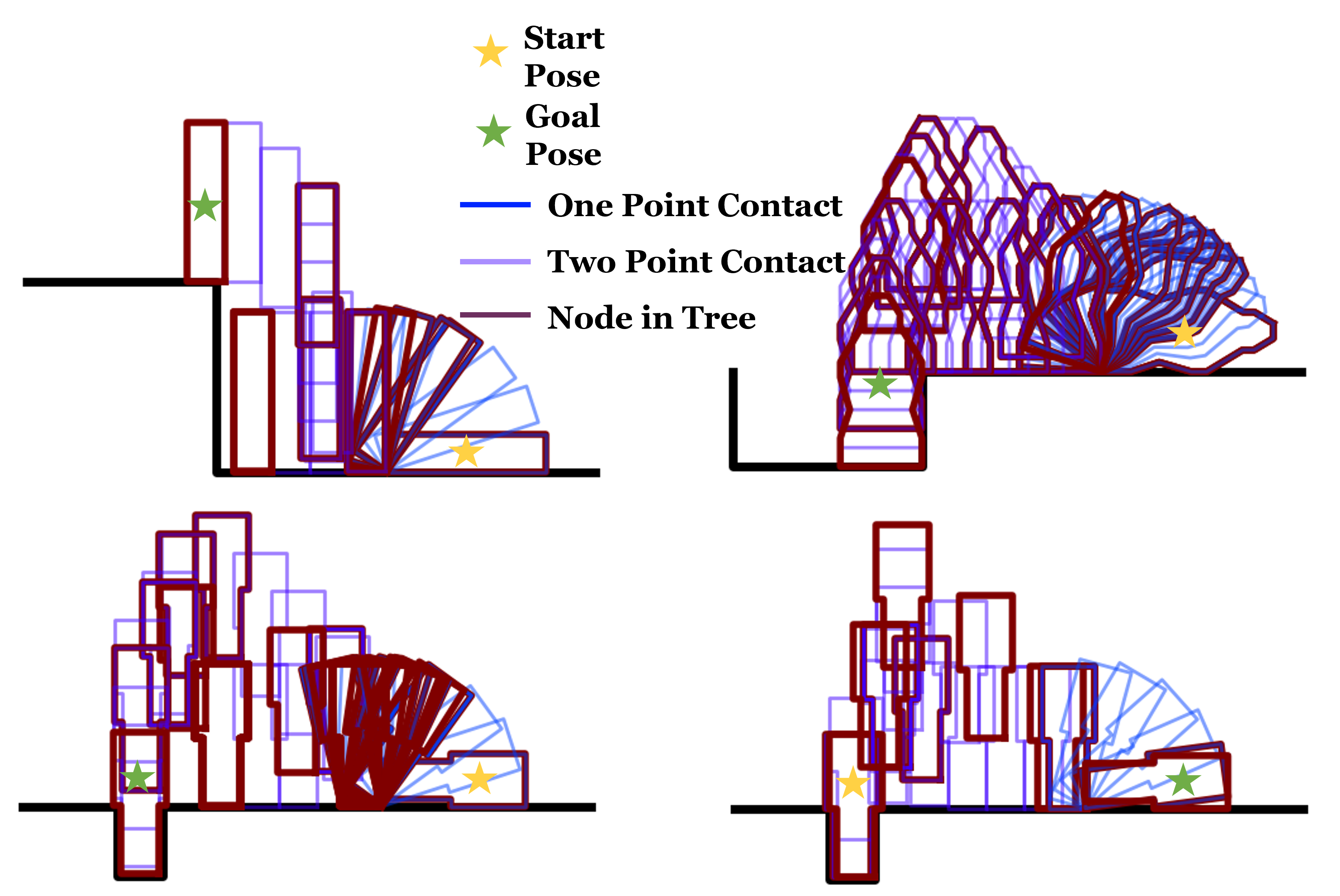}
    \makebox[1.0\linewidth]{\footnotesize (b) Trees }
    \caption{Plans generated by the multi-modal manipulation planner for four tasks. (a) Waypoint object poses along solution trajectories, with colors representing different manipulation modes. (b) Trees explored by the planner corresponding to the above trajectories. Nodes are highlighted in bold red and waypoints along edges are colored in the same manner as above. [Best viewed in color.]}
    \label{fig:trajs}
\end{figure}

Results on the 3 tasks of Fig.~\ref{fig:tasks} are illustrated in Fig.~\ref{fig:trajs}.  The solution trajectories demonstrate that the planner discovers the changes of manipulation mode from one point contact to two point contact or vice versa, and can switch from pivoting to grasping to sliding.  The sampled trees are also plotted, illustrating that very few nodes are actually sampled and the planner makes quite direct progress toward the goal. 

Timing and success rates on the same tasks as well as their {\em reversed} versions, in which the start and goal pose are interchanged, are shown in Tab.~\ref{tab:2}
.  We see that STOCS is only called a few dozen times at most, and the transition test is effective at rejecting ineffective pose samples.  We also explored how the planner performs  as the total number of time steps $T$ in STOCS is varied. As can be seen in Tab.~\ref{tab:5}, the success rate slightly increases as $T$ increases and the number of nodes in the tree and solution path  decreases. This can be explained by the longer time horizon enabling STOCS to make larger steps in the state space, giving a better chance to connect to the goal pose within the sample limit.  However, a larger $T$ increases solve times overall, since each call of the local planner solves a larger optimization problem.

\begin{table}[]
\renewcommand{\arraystretch}{1.2}
\setlength\tabcolsep{3pt}
\centering
\begin{tabular}{@{}llcccccc@{}}
\toprule
\multicolumn{2}{l}{Direction}                                                                      & \multicolumn{3}{c}{Forward} & \multicolumn{3}{c}{Reverse} \\ 
\multicolumn{2}{l}{Task}                                                                           & 1       & 2       & 3       & 1       & 2       & 3       \\ \hline
\multicolumn{2}{l}{Success} & 9/10    & 9/10    & 8/10    & 10/10   & 10/10   & 10/10   \\ \hline
\multirow{2}{*}{\begin{tabular}[c]{l}Nodes\\ (median)\end{tabular}}         & in tree        & 26      & 21      & 18      & 8       & 14      & 10      \\ 
                                                                                  & in path        & 13      & 14      & 13      & 7       & 11      & 6       \\ \hline
\multirow{3}{*}{Time (s)}                                                         & min            & 334     & 633     & 401     & 99      & 327     & 316     \\ 
                                                                                  & median         & 1300    & 1310    & 1048    & 425     & 3360    & 2014    \\ 
                                                                                  & max            & 3712    & 4472    & 1837    & 3747    & 5882    & 7712    \\ \hline
\multicolumn{2}{l}{\begin{tabular}[c]{@{}l@{}}STOCS calls\\ (median)\end{tabular}}                & 25      & 24      & 27      & 7       & 15      & 10      \\ \bottomrule
\end{tabular}
\caption{ Success rate, number of nodes in the tree and the path, the planning time, and the number of STOCS called of the proposed planner on 3 tasks with different initial and goal pose. Forward direction has the same start and goal pose as shown in \prettyref{fig:tasks}, and reverse direction has the start and goal pose interchanged. \label{tab:2}}
\end{table}

\begin{table}[]\renewcommand{\arraystretch}{1.2}
\setlength\tabcolsep{1pt}
\centering
\begin{tabular}{@{}llccccccccc@{}}
\toprule
\multicolumn{2}{l}{T}                                    & \multicolumn{3}{c}{3} & \multicolumn{3}{c}{5} & \multicolumn{3}{c}{10} \\ \hline
\multicolumn{2}{l}{Task}                                 & 1      & 2      & 3      & 1       & 2      & 3      & 1      & 2      & 3     \\ \hline
\multicolumn{2}{l}{Success} & 9/10   & 8/10   & 8/10   & 9/10    & 9/10   & 8/10   & 8/10   & 10/10  & 10/10      \\ \hline
\multirow{2}{*}{\begin{tabular}[l]{l}Nodes\\ (median)\end{tabular}}         & in tree        & 39     & 28     & 18     & 26      & 21     & 18     & 18     & 12     & 6      \\ 
                                        & in path        & 14     & 20     & 13     & 13      & 14     & 13     & 12     & 9      & 5      \\ \hline
\multirow{3}{*}{Time (s)}             & min            & 422    & 410    & 221    & 334     & 633    & 401    & 595    & 926    & 606     \\ 
                                        & median         & 915    & 906    & 385    & 1300    & 1310   & 1048   & 2072   & 1726   & 1648   \\  
                                        & max            & 1884   & 1752   & 1275   & 3712    & 4472   & 1837   & 6994   & 2959   &  5590  \\ \bottomrule
\end{tabular}
\caption{ Success rate, number of nodes in the tree and the path, and the planning time of the proposed planner on 3 tasks with different total number of time steps $T$ for STOCS. \label{tab:5}}
\end{table}

Hardware experiments are performed to evaluate the planned trajectories. A Mitsubishi Electric Assista industrial position-controlled arm with a F/T sensor mounted at the wrist of the robot is used in the experiment. The default stiffness controller of the robot is used to execute the planned force trajectories. To implement the optimal force trajectory on the object, we design a reference trajectory for the robot that presses into the object such that the robot would apply the desired force for the estimated stiffness constants for the low-level robot position control.

\begin{figure}
    \centering
    \includegraphics[width=0.45\textwidth]{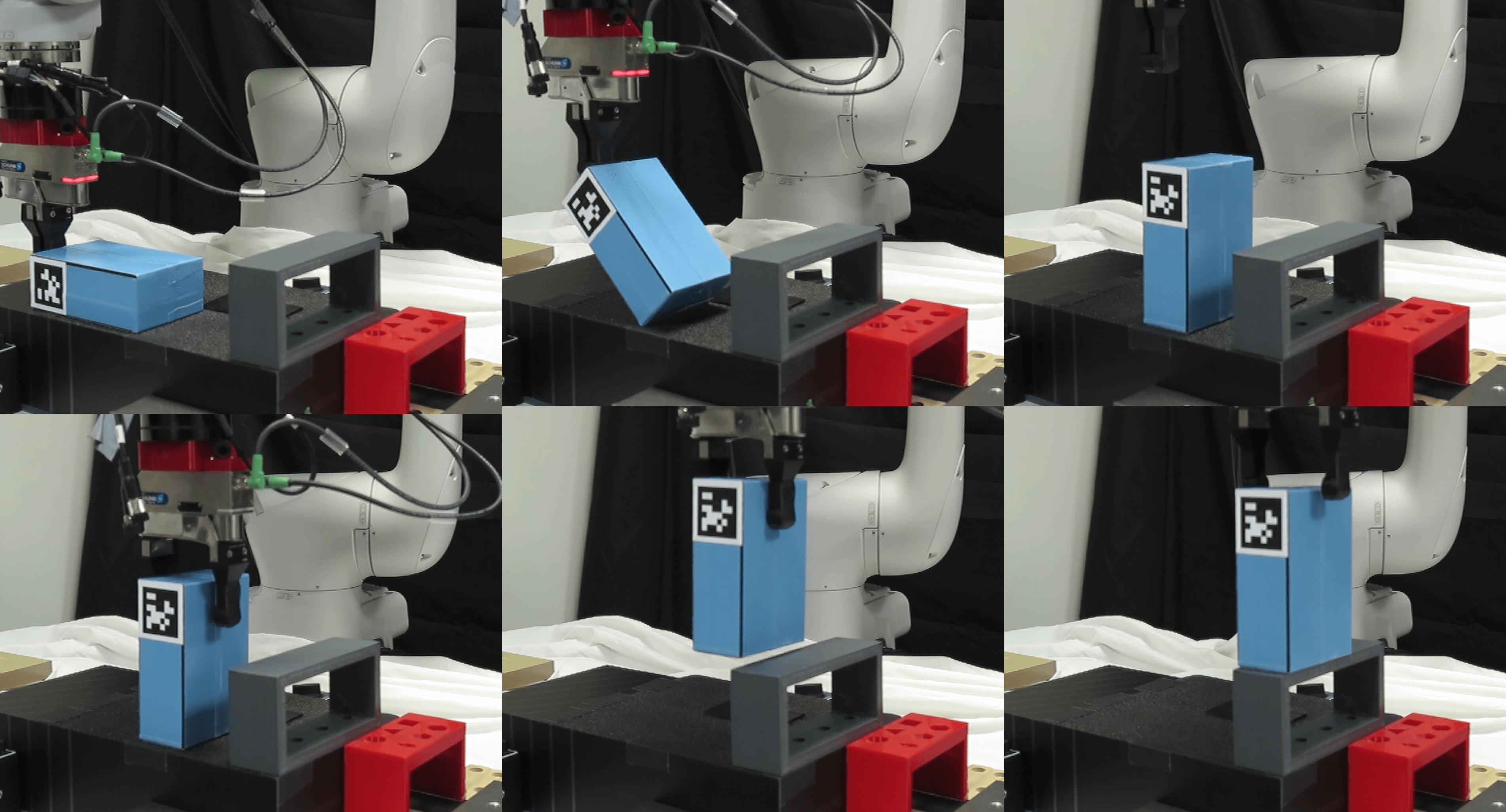}
    \makebox[1.0\linewidth]{\footnotesize (a) Reorient and place a box}\vspace{2mm}
    \includegraphics[width=0.45\textwidth]{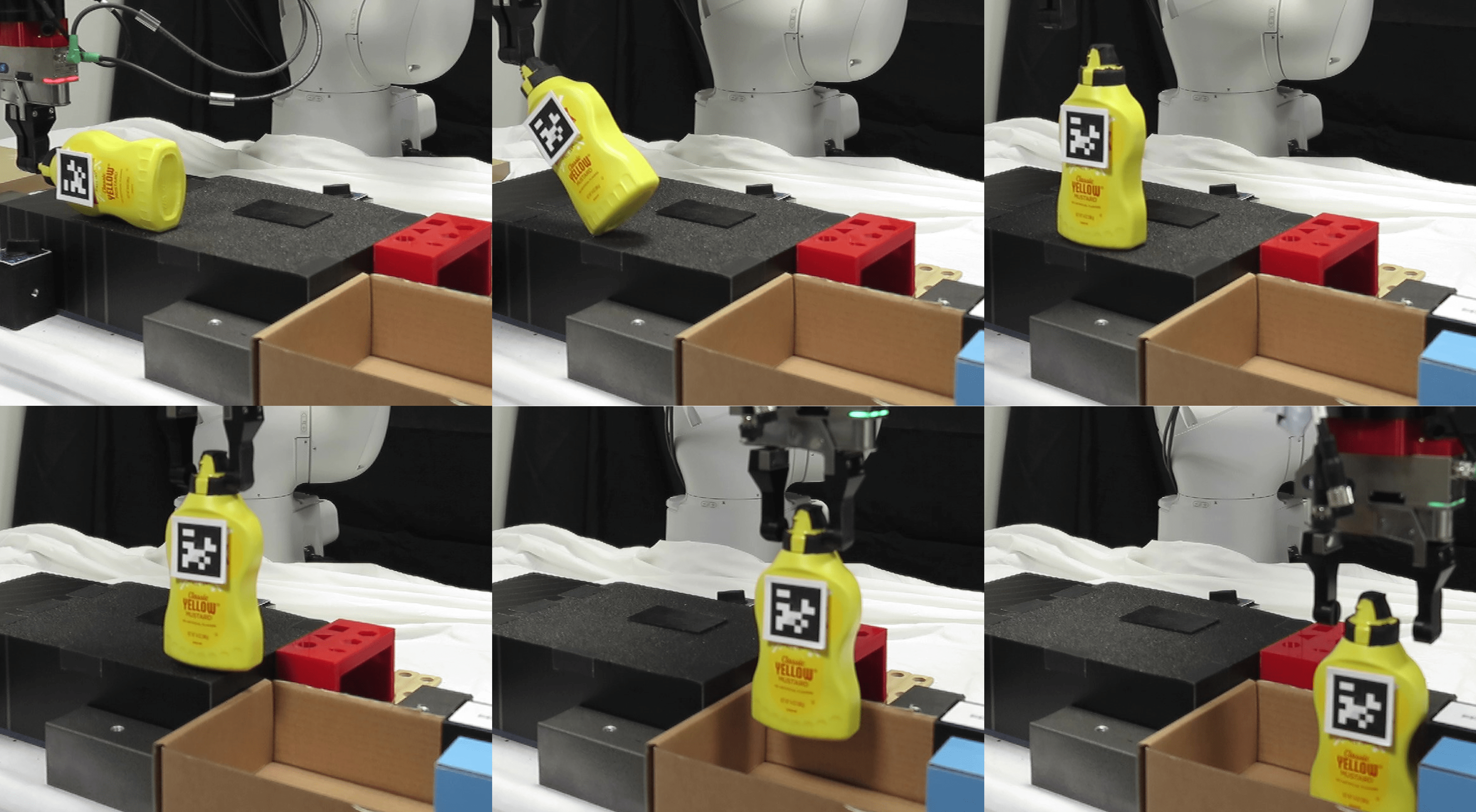}
    \makebox[1.0\linewidth]{\footnotesize (b) Pack a mustard bottle}\vspace{2mm}
    \includegraphics[width=0.45\textwidth]{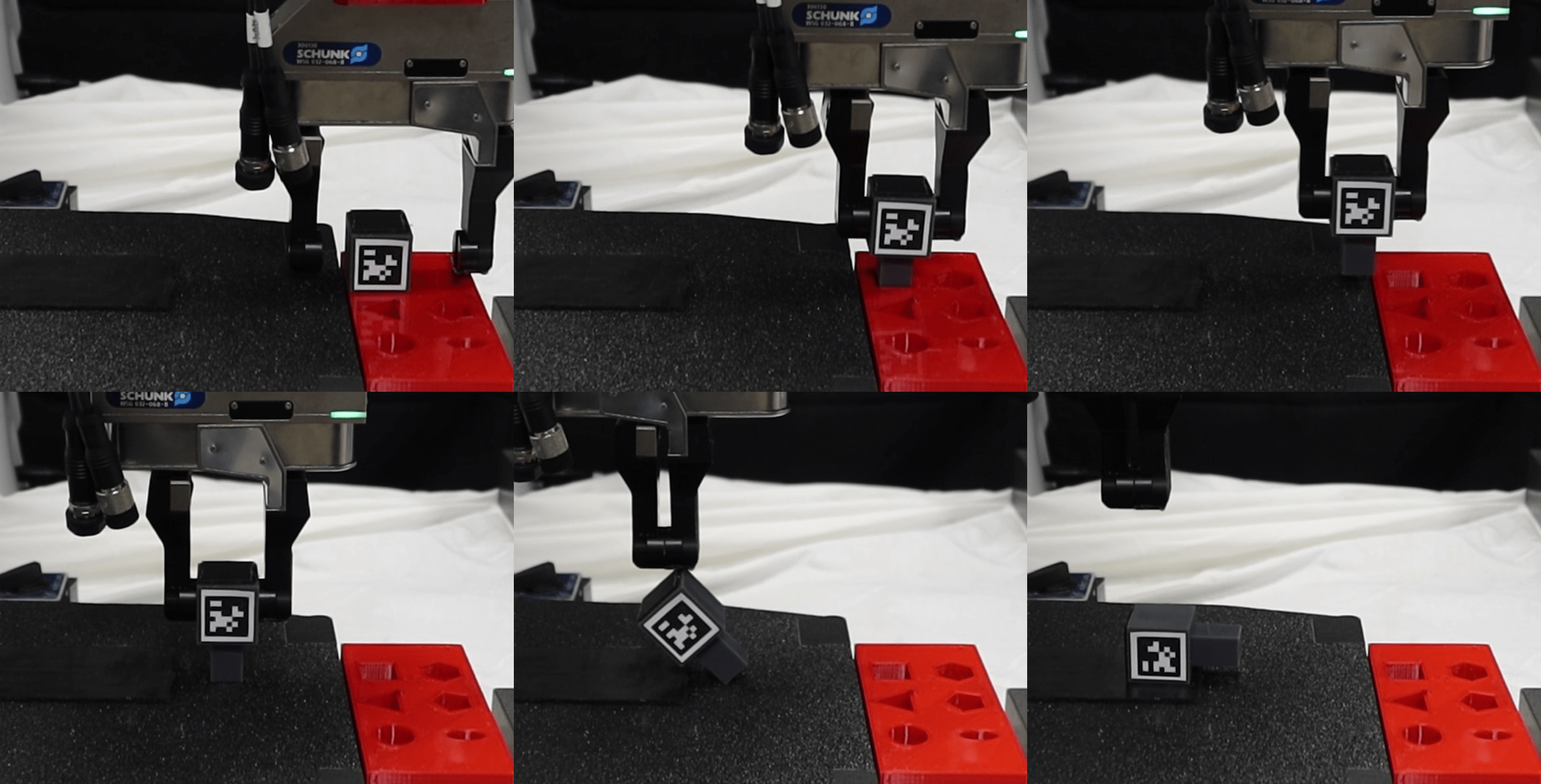}
    \makebox[1.0\linewidth]{\footnotesize (c) Unplug and lay down a peg}
    \caption{Snapshots along the trajectories executed on the real robot. (a) Reorient and place a box. (b) Pack a mustard bottle. (c) Unplug a peg and lay it down on a plane. AprilTag pose feedback is used to adjust trajectories only when the manipulator contact state changes.  [Best viewed in color.]}
    \label{fig:hardware_trajs}
\end{figure}

Since the computed trajectory is executed without object pose feedback, execution error can accumulate. Thus, we use AprilTags \cite{olson2011apriltag} (Fig.~\ref{fig:hardware_trajs}) to track the pose of the object, and after a single-mode manipulation trajectory is completed, the object pose feedback is used to adjust the execution of the next mode's trajectory.  Some trajectories recorded during the robot experiments are shown in Fig.~\ref{fig:hardware_trajs}.

\section{Conclusion and Discussion}\label{sec:conclusion}
This paper proposed STOCS, a novel contact-implicit trajectory optimization and infinite programming algorithm to generate manipulation trajectories involving sliding and pivoting. It also proposed the use of STOCS in a multi-modal manipulation planner that uses sampling-based planning to generate plans involving changes of manipulator contact state. Experiments show that STOCS scales to complex geometries better than standard contact-invariant optimization, and the proposed multi-modal planner is validated on several manipulation tasks in simulation and on a real robot.

One limitation of STOCS is that it assumes quasi-static motion. We plan to investigate how the quasi-dynamic assumption, which has been used in previous work \cite{chavan2020planar,cheng2022contact,pang2022global}, can be formulated in an infinite programming framework. We also plan to investigate speeding up the optimization, possibly using sequential quadratic programming \cite{gill2005snopt} with warm starts and incorporating time-active contact sets and deactivating contacts to decrease the number of variables and constraints.

Future work could also refine the multi-modal planner. Experiments suggest that approximately half or more of the time spent in STOCS fails to make progress toward the target object pose, since the selected robot contact points cannot manipulate the object in the desired direction (for example, the second row of Fig.~\ref{fig:stocs_examples}). Methods for sampling manipulator modes so that the target configuration is reachable (and vice versa) would reduce computation times significantly.  We also wish to explore the use of more sophisticated grasp generators in manipulator mode sampling, and explore applying our algorithms to fully 3D problems.

Finally, we note that our planned paths may be susceptible to pose or environmental uncertainty during execution.  Possible approaches may include robust optimization techniques~\cite{9811812}, or feedback controllers to monitor the contact state and adjust the control appropriately~\cite{hogan2020tactile}. We plan to explore these avenues in future work.

\section*{Acknowledgments}
This paper is partially supported by NSF Grant \#IIS-1911087.

\bibliographystyle{plainnat}
\bibliography{references}

\begin{thebibliography}{28}
\providecommand{\natexlab}[1]{#1}
\providecommand{\url}[1]{\texttt{#1}}
\expandafter\ifx\csname urlstyle\endcsname\relax
  \providecommand{\doi}[1]{doi: #1}\else
  \providecommand{\doi}{doi: \begingroup \urlstyle{rm}\Url}\fi

\bibitem[Anderson and Philpott(2012)]{anderson2012infinite}
Edward~J Anderson and Andrew~B Philpott.
\newblock \emph{Infinite Programming: Proceedings of an International Symposium
  on Infinite Dimensional Linear Programming Churchill College, Cambridge,
  United Kingdom, September 7--10, 1984}, volume 259.
\newblock Springer Science \& Business Media, 2012.

\bibitem[Chavan-Dafle and Rodriguez(2020)]{chavan2020sampling}
Nikhil Chavan-Dafle and Alberto Rodriguez.
\newblock Sampling-based planning of in-hand manipulation with external pushes.
\newblock In \emph{Robotics Research: The 18th International Symposium ISRR},
  pages 523--539. Springer, 2020.

\bibitem[Chavan-Dafle et~al.(2020)Chavan-Dafle, Holladay, and
  Rodriguez]{chavan2020planar}
Nikhil Chavan-Dafle, Rachel Holladay, and Alberto Rodriguez.
\newblock Planar in-hand manipulation via motion cones.
\newblock \emph{The International Journal of Robotics Research}, 39\penalty0
  (2-3):\penalty0 163--182, 2020.

\bibitem[Cheng et~al.(2021)Cheng, Huang, Hou, and Mason]{cheng2021contact}
Xianyi Cheng, Eric Huang, Yifan Hou, and Matthew~T Mason.
\newblock Contact mode guided sampling-based planning for quasistatic dexterous
  manipulation in 2d.
\newblock In \emph{2021 IEEE International Conference on Robotics and
  Automation (ICRA)}, pages 6520--6526. IEEE, 2021.

\bibitem[Cheng et~al.(2022)Cheng, Huang, Hou, and Mason]{cheng2022contact}
Xianyi Cheng, Eric Huang, Yifan Hou, and Matthew~T Mason.
\newblock Contact mode guided motion planning for quasidynamic dexterous
  manipulation in 3d.
\newblock In \emph{2022 International Conference on Robotics and Automation
  (ICRA)}, pages 2730--2736. IEEE, 2022.

\bibitem[Dikin(1967)]{dikin1967iterative}
I.I. Dikin.
\newblock Iterative solution of problems of linear and quadratic programming.
\newblock In \emph{Doklady Akademii Nauk}, volume 174, pages 747--748. Russian
  Academy of Sciences, 1967.

\bibitem[Garrett et~al.(2021)Garrett, Chitnis, Holladay, Kim, Silver,
  Kaelbling, and Lozano-P{\'e}rez]{garrett2021integrated}
Caelan~Reed Garrett, Rohan Chitnis, Rachel Holladay, Beomjoon Kim, Tom Silver,
  Leslie~Pack Kaelbling, and Tom{\'a}s Lozano-P{\'e}rez.
\newblock Integrated task and motion planning.
\newblock \emph{Annual Review of Control, Robotics, and Autonomous Systems},
  4:\penalty0 265--293, 2021.

\bibitem[Gill et~al.(2005)Gill, Murray, and Saunders]{gill2005snopt}
Philip~E Gill, Walter Murray, and Michael~A Saunders.
\newblock Snopt: An sqp algorithm for large-scale constrained optimization.
\newblock \emph{SIAM review}, 47\penalty0 (1):\penalty0 99--131, 2005.

\bibitem[Harada et~al.(2006)Harada, Hauser, Bretl, and
  Latombe]{harada2006natural}
Kensuke Harada, Kris Hauser, Tim Bretl, and Jean-Claude Latombe.
\newblock Natural motion generation for humanoid robots.
\newblock In \emph{2006 IEEE/RSJ International Conference on Intelligent Robots
  and Systems}, pages 833--839. IEEE, 2006.

\bibitem[Hauser(2021)]{hauser2021semi}
Kris Hauser.
\newblock Semi-infinite programming for trajectory optimization with non-convex
  obstacles.
\newblock \emph{The International Journal of Robotics Research}, 40\penalty0
  (10-11):\penalty0 1106--1122, 2021.

\bibitem[Hogan et~al.(2020)Hogan, Ballester, Dong, and
  Rodriguez]{hogan2020tactile}
Francois~R Hogan, Jose Ballester, Siyuan Dong, and Alberto Rodriguez.
\newblock Tactile dexterity: Manipulation primitives with tactile feedback.
\newblock In \emph{2020 IEEE international conference on robotics and
  automation (ICRA)}, pages 8863--8869. IEEE, 2020.

\bibitem[Huang et~al.(2021)Huang, Cheng, and Mason]{huang2021efficient}
Eric Huang, Xianyi Cheng, and Matthew~T Mason.
\newblock Efficient contact mode enumeration in 3d.
\newblock In \emph{International Workshop on the Algorithmic Foundations of
  Robotics}, pages 485--501. Springer, 2021.

\bibitem[Jaillet et~al.(2010)Jaillet, Cort{\'e}s, and
  Sim{\'e}on]{jaillet2010sampling}
L{\'e}onard Jaillet, Juan Cort{\'e}s, and Thierry Sim{\'e}on.
\newblock Sampling-based path planning on configuration-space costmaps.
\newblock \emph{IEEE Transactions on Robotics}, 26\penalty0 (4):\penalty0
  635--646, 2010.

\bibitem[LaValle et~al.(1998)]{lavalle1998rapidly}
Steven~M LaValle et~al.
\newblock Rapidly-exploring random trees: A new tool for path planning.
\newblock 1998.

\bibitem[L{\'o}pez and Still(2007)]{lopez2007semi}
Marco L{\'o}pez and Georg Still.
\newblock Semi-infinite programming.
\newblock \emph{European Journal of Operational Research}, 180\penalty0
  (2):\penalty0 491--518, 2007.

\bibitem[Manchester et~al.(2019)Manchester, Doshi, Wood, and
  Kuindersma]{manchester2019contact}
Zachary Manchester, Neel Doshi, Robert~J Wood, and Scott Kuindersma.
\newblock Contact-implicit trajectory optimization using variational
  integrators.
\newblock \emph{The International Journal of Robotics Research}, 38\penalty0
  (12-13):\penalty0 1463--1476, 2019.

\bibitem[Mason(2018)]{mason2018toward}
Matthew~T Mason.
\newblock Toward robotic manipulation.
\newblock \emph{Annual Review of Control, Robotics, and Autonomous Systems},
  1\penalty0 (1), 2018.

\bibitem[Mordatch et~al.(2012{\natexlab{a}})Mordatch, Popovi{\'c}, and
  Todorov]{mordatch2012contact}
Igor Mordatch, Zoran Popovi{\'c}, and Emanuel Todorov.
\newblock Contact-invariant optimization for hand manipulation.
\newblock In \emph{Proceedings of the ACM SIGGRAPH/Eurographics Symposium on
  Computer Animation}, pages 137--144, 2012{\natexlab{a}}.

\bibitem[Mordatch et~al.(2012{\natexlab{b}})Mordatch, Todorov, and
  Popovi{\'c}]{mordatch2012discovery}
Igor Mordatch, Emanuel Todorov, and Zoran Popovi{\'c}.
\newblock Discovery of complex behaviors through contact-invariant
  optimization.
\newblock \emph{ACM Transactions on Graphics (TOG)}, 31\penalty0 (4):\penalty0
  1--8, 2012{\natexlab{b}}.

\bibitem[Olson(2011)]{olson2011apriltag}
Edwin Olson.
\newblock Apriltag: A robust and flexible visual fiducial system.
\newblock In \emph{2011 IEEE international conference on robotics and
  automation}, pages 3400--3407. IEEE, 2011.

\bibitem[Pang et~al.(2022)Pang, Suh, Yang, and Tedrake]{pang2022global}
Tao Pang, HJ~Suh, Lujie Yang, and Russ Tedrake.
\newblock Global planning for contact-rich manipulation via local smoothing of
  quasi-dynamic contact models.
\newblock \emph{arXiv preprint arXiv:2206.10787}, 2022.

\bibitem[Posa et~al.(2014)Posa, Cantu, and Tedrake]{posa2014direct}
Michael Posa, Cecilia Cantu, and Russ Tedrake.
\newblock A direct method for trajectory optimization of rigid bodies through
  contact.
\newblock \emph{The International Journal of Robotics Research}, 33\penalty0
  (1):\penalty0 69--81, 2014.

\bibitem[Raghunathan et~al.(2022)Raghunathan, Jha, and
  Romeres]{raghunathan2022pyrobocop}
Arvind~U Raghunathan, Devesh~K Jha, and Diego Romeres.
\newblock Pyrobocop: Python-based robotic control \& optimization package for
  manipulation.
\newblock In \emph{2022 International Conference on Robotics and Automation
  (ICRA)}, pages 985--991. IEEE, 2022.

\bibitem[Reemtsen and G{\"o}rner(1998)]{reemtsen1998numerical}
Rembert Reemtsen and Stephan G{\"o}rner.
\newblock Numerical methods for semi-infinite programming: a survey.
\newblock In \emph{Semi-Infinite Programming}, pages 195--275. Springer, 1998.

\bibitem[Shirai et~al.(2022)Shirai, Jha, Raghunathan, and Romeres]{9811812}
Yuki Shirai, Devesh~K. Jha, Arvind~U. Raghunathan, and Diego Romeres.
\newblock Robust pivoting: Exploiting frictional stability using bilevel
  optimization.
\newblock In \emph{2022 International Conference on Robotics and Automation
  (ICRA)}, pages 992--998. IEEE, 2022.

\bibitem[Toussaint et~al.(2018)Toussaint, Allen, Smith, and
  Tenenbaum]{toussaint2018differentiable}
Marc~A Toussaint, Kelsey~Rebecca Allen, Kevin~A Smith, and Joshua~B Tenenbaum.
\newblock Differentiable physics and stable modes for tool-use and manipulation
  planning.
\newblock In \emph{Robotics: Science and Systems}, 2018.

\bibitem[Winkler et~al.(2018)Winkler, Bellicoso, Hutter, and
  Buchli]{winkler2018gait}
Alexander~W Winkler, C~Dario Bellicoso, Marco Hutter, and Jonas Buchli.
\newblock Gait and trajectory optimization for legged systems through
  phase-based end-effector parameterization.
\newblock \emph{IEEE Robotics and Automation Letters}, 3\penalty0 (3):\penalty0
  1560--1567, 2018.

\bibitem[Zhang and Hauser(2021)]{zhang2021semi}
Mengchao Zhang and Kris Hauser.
\newblock Semi-infinite programming with complementarity constraints for pose
  optimization with pervasive contact.
\newblock In \emph{2021 IEEE International Conference on Robotics and
  Automation (ICRA)}, pages 6329--6335. IEEE, 2021.

\end{thebibliography}

\newpage
\end{document}